\documentclass[lettersize,journal]{IEEEtran}
\usepackage{amsmath,amsfonts}
\usepackage{array}
\usepackage[caption=false,font=normalsize,labelfont=sf,textfont=sf]{subfig}
\usepackage{textcomp}
\usepackage{stfloats}
\usepackage{url}
\usepackage{verbatim}
\usepackage{graphicx}
\usepackage{subcaption}

\usepackage{graphicx}   

\usepackage{amsmath,amssymb,amsfonts}

\usepackage{amsthm}
\usepackage{hyperref}
\usepackage{mathrsfs}
\usepackage{enumitem}

\usepackage{booktabs}
\usepackage{multirow}
\usepackage{boldline} 

\usepackage{algpseudocode} 
\usepackage[linesnumbered,ruled]{algorithm2e}
\usepackage{bm}

\newcommand*{\modelname}{\textsf{MNDE}\@\xspace}
\usepackage{tikz}
\usepackage{cancel}

\usepackage{cite}
\begin{document}

\title{Multi-View Neural Differential Equations for Continuous-Time Stream Data in Long-Term Traffic Forecasting}

\author{Zibo Liu, Zhe Jiang,~\IEEEmembership{Senior Member,~IEEE,} and Shigang Chen,~\IEEEmembership{Fellow,~IEEE} % stops a space
\thanks{Zibo Liu, Zhe Jiang, and Shigang Chen are with the Department of Computer Information Science \& Engineering, University of Florida, Gainesville, FL 32611 USA. (Email: ziboliu@ufl.edu, zhe.jiang@ufl.edu, sgchen@cise.ufl.edu)}}% <-this % stops a space



\maketitle

\begin{abstract}
Long-term traffic flow forecasting plays a crucial role in intelligent transportation as it allows traffic managers to adjust their decisions in advance. However, the problem is challenging due to spatio-temporal correlations and complex dynamic patterns in continuous-time stream data. Neural Differential Equations (NDEs) are among the state-of-the-art methods for learning continuous-time traffic dynamics. However, the traditional NDE models face issues in long-term traffic forecasting due to failures in capturing delayed traffic patterns,  dynamic edge (location-to-location correlation) patterns, and abrupt trend patterns. To fill this gap, we propose a new NDE architecture called Multi-View Neural Differential Equations. Our model captures current states, delayed states, and trends in different state variables (views) by learning latent multiple representations within Neural Differential Equations. Extensive experiments conducted on several real-world traffic datasets demonstrate that our proposed method outperforms the state-of-the-art and achieves superior prediction accuracy for long-term forecasting and robustness with noisy or missing inputs. 
\end{abstract}


\section{Introduction}\label{sec:introduction}

\IEEEPARstart{G}{iven} historical traffic flow measurements at a selected set of locations in a road network, long-term spatio-temperal traffic forecasting aims to train a deep-learning model to predict the traffic flow at those locations far ahead into the future.  Consider a scenario in a road network where the long-term forecasting model could utilize one hour of recent past data to predict the traffic conditions up to eight hours ahead, where the amount of future predicted data is eight times the past measurements. Whereas the short-term forecasting model could only predict one hour of traffic data ahead, which is far less than what the long-term forecasting model did.  The capability of long-term traffic forecasting can play a crucial role in intelligent transportation. For example, early warnings about future traffic conditions may help traffic planners (such as map apps), traffic signal management, and agencies of law enforcement, medical assistance and disaster response to adjust their decisions and routes \cite{cheng2020mitigating,balasubramanian2015adaptive}.

Existing methods for traffic flow forecasting can be categorized into discrete-time or continuous-time methods. Discrete-time methods include recurrent neural networks (RNNs) \cite{veeriah2015differential,NIPS2015_07563a3f,Zhang_Zheng_Qi_2017}, or convolutional neural networks (CNNs) with graph neural networks (GNNs) \cite{yu2017spatio,li2018diffusion,wu2019graph,li2020spatial,Jiang_Wang_Yong_Jeph_Chen_Kobayashi_Song_Fukushima_Suzumura_2023,ji2023spatio}. These methods cannot fully capture the dynamic patterns of traffic flows in continuous time, especially for discrete traffic measurements with noise and missing values. In addition, GNNs are shown to contain the risk of over-smoothing when trained on limited data \cite{graph-over-smoothing1,graph-over-smoothing2}. Continuous-time methods are largely based on neural differential equations (NDEs) \cite{chen2018neuralode,schirmer2022modeling,kidger2020neural,kidger2022neural}. These methods train a neural network to capture the continuous-time traffic dynamics expressed by a differential equation \cite{fang2021spatial,choi2022graph,ji2022stden,jin2022multivariate}. Specifically, a neural network is trained to fit the gradient function. However, existing NDE models do not fully capture some complex dynamic patterns such as delayed propagation of traffic conditions, dynamic spatial dependency in traffic dynnamics, and abrupt shifts in local traffic trend, which are elaborated below.  

First, their designs lack consideration of the complex cascading effects with delayed propagation of traffic conditions that are typical in traffic flows. For example, consider a scenario where a slow-moving vehicle on a highway prompts deceleration, which triggers a cascade effect as the following drivers also reduce speed with a delay. The time, location, and extent of this delayed pattern are unknown beforehand. Second, the spatial dependency between locations in a road network is temporally dynamic due to rush hours and accidents. How to {\it explicitly} and {\it quantitatively} capture such dependency remains a challenge. Third, there exist abrupt shifts in traffic flow patterns that are local and deviate from the current broad traffic conditions, due to certain events such as sudden road closures caused by car accidents or bad weather. This requires us to incorporate additional input on local traffic trend which is missing in the existing spatio-temporal models.  
The impact of considering delayed propagation, treating spatial dependency explicitly, and incorporating local traffic trend is less significant on short-term traffic forecasting, which most existing work focuses on \cite{fan2022depts,wu2019graph,yu2017spatio,song2020spatial,li2020spatial,Jiang_Wang_Yong_Jeph_Chen_Kobayashi_Song_Fukushima_Suzumura_2023,chen2021z,fang2021spatial,choi2022graph,yi2023fouriergnn,ji2023spatio}, but is more significant on long-term traffic forecasting, which this work focuses on.

We propose a new deep-learning framework called  \textbf{M}ulti-View \textbf{N}eural \textbf{D}ifferential \textbf{E}quations (\textbf{\modelname}) for long-term spatio-temporal traffic forecasting. It uses NDEs as the backbone to capture dynamic traffic patterns in continuous time. More importantly, our framework contains separate NDE modules to capture the complex delayed propagation and dynamic spatial dependency. Specifically, our model incorporates dynamic spatial correlations by calculating intermediate correlations between locations over time as a new NDE input. Finally, we model abrupt shifts in traffic dynamics through a differentiation module, which conducts self-attention on the gradients of traffic flow in continuous time. The proposed framework uses multiple neural differential equations, which offer different views into the complex spatio-temporal dependencies in traffic flow data. Extensive experiments show that our framework outperforms the state-of-the-art models for long-term traffic forecasting in its accuracy, especially when the input traffic flow data are noisy.

\section{Preliminaries and Problem Statement}
\label{sec:related_works}

\subsection{Preliminaries}

\noindent{\itshape Definition 1: Flow Rate Measurements.}\label{Interpolate}
A road system has a set $\mathcal{N}$ of chosen locations where sensors are deployed to measure traffic. Denote $n = |\mathcal{N}|$. 
All the vehicles that pass a location form a traffic flow; a sensor at the location measures the {\it flow rate}, i.e., the number of passing vehicles at the location during each preset time interval. Let $\mathcal{X}= [\mathcal{X}_{i,j}, i \in \mathcal{N}, j \in \{0,1,..,l-1\} ]^T$ be the flow-rate measurement matrix across all locations over $l$ consecutive time intervals, where $\mathcal{X}_{i,j}$ is the flow-rate measurement at location $i$ during the $j$th time interval. For simplicity, we normalize each time interval as one unit of time. Denote the $l$ measurements at location $i$ as $\mathcal{X}_i = [\mathcal{X}_{i,j}, j \in \{0,1,..,l-1\}]^T$. Hence, $\mathcal{X}=[\mathcal{X}_i, i \in \mathcal{N} ]^T$.

To facilitate the construction of neural differential equations, we need to interpolate the discrete flow-rate measurements $\mathcal{X}_i$ at each location $i$ to a continuous time function $X_i(t)$, $t \in [0, l-1]$, with $X_i(j) = {X}_{i, j}$, $\forall j \in \{0,1,..,l-1\}$. One common method is the natural cubic spline method \cite{mckinley1998cubic, kidger2020neural} that produces a segmented curve: $X_{i}(t) = a_{i,j} + b_{i,j} \left(t-t_j\right) + c_{i,j}\left(t-t_j\right)^2 + d_{i,j}\left(t-t_j\right)^3$, $\forall t \in[t_j,t_{j+1}]$, $j \in \{0,1,..,l-2\}$, where $a_{i,j}$, $b_{i,j}$, $c_{i,j}$, $d_{i,j}$ are interpolation coefficients. Those coefficients can be computed based on the spline interpolation condition: $X_i(t=j) = \mathcal{X}_{i,j}$, the continuity condition: $X_{i}(t= j+1) = \mathcal{X}_{i,j+1}$, and the second derivative condition: $X_{i}^{\prime \prime}(t=j)=X_{i}^{\prime \prime}(t=t_j+1)=0$. We denote $X(t)=[X_i(t), i \in \mathcal{N} ]^T$ as a vector of flow-rate functions at all locations. We sometimes abbreviate $X(t)$ as $X$ for simplicity.

\noindent{\itshape Definition 2: Neural Differential Equation (NDE) \cite{chen2018neural,DBLP:journals/corr/abs-2202-02435}.}
% \subsection{Neural Differential Equation (NDE)}
Let's begin with the Neural Ordinary Differential Equation (NODE) \cite{fang2021spatial,node_rnn1,lechner2020learningy,liu2023graphbased}. Consider a latent variable, which is the embedding  of $X$ generated from a fully connected (FC) layer, denoted as $H = [H_i, i \in \mathcal{N}]^T$. 
\begin{align}
H(t) &=H(0)+\int_{0}^{t} H'(\tau) \mathrm{d} \tau 
\end{align}
One can build a NODE neural network, denoted as $f_o(H(t): \theta_o)$, which takes $H(t)$ as input and produces an approximate value for gradient function $H'(t)$, where $\theta_o$ denotes the parameters, which can be optimized through training. 
\begin{align}
H(t) &\approx H(0)+\int_{0}^{t} f_o(H(\tau): \theta_o) \mathrm{d} \tau
\label{NODE}
\end{align}
By approximating the integral with a summation over small steps and iteratively applying $f_o$, we can compute $H(t)$ iteratively, starting from $H(0)$, through an ODE solver \cite{chen2018neural}.

It is shown that the controlled version of NODE, Neural Controlled Differential Equation (NCDE) \cite{NEURIPS2020_4a5876b4,ncde_online}, generates more accurate results. It employs a neural network $f_c(H(t): \theta_c)$, taking $H(t)$ as input and producing an output, which is then multiplied by the control term $X'(t)$ to approximate $H'(t)$, where $\theta_c$ denotes the parameters. 
%The control term enables more flexibility in capturing temporal patterns. 
\begin{align}
\label{NCDE:1}
H(t) &\approx H(0)+\int_{0}^{t} f_c(H(\tau): \theta_c)  X'(\tau)\mathrm{d} \tau
\end{align}
The integral in (\ref{NCDE:1}) is a Riemann–Stieltjes integral problem \cite{dragomir1998inequality} whereas that in (\ref{NODE}) is a Riemann integral problem.

\subsection{Problem Statement}

Design a deep learning network, denoted as MNDE (Multi-view NDE), which takes the past measurements over $l$ time intervals as input and produces forcast on the flow rates of the next $l'$ time intervals. The number of history intervals $l$ is far less than the number of future intervals $l'$.

In most work \cite{fang2021spatial,song2020spatial,jin2022automated,shao2022spatial,jia2020residual,ji2022stden}, both $l$ and $l'$ are set to 1 hour for short-term forecast. Note that $l$ should not be too large to avoid excessively large models and computation costs that come with them. Moreover, research has shown that too large $l$ may actually degrade forecast accuracy \cite{zeng2023transformers}. For long-term forecast, $l'$ is set much larger than $l$. 

\noindent{\textbf{Input:}}

\noindent{$\bullet$} $\mathcal{X}= [\mathcal{X}_{i,j}, i \in \mathcal{N}, j \in \{0,1,..,l-1\} ]^T$ as the past measurements over $l$ time intervals

\noindent{$\bullet$} $\mathcal{Y}= [\mathcal{Y}_{i,j}, i \in \mathcal{N}, j \in \{0,1,..,l'-1\} ]^T$ as the ground truth on the flow rates measured over the next $l'$ intervals

%\noindent{$\bullet$} Different type of Differential Equations and their functions for traffic dynamics.

\noindent{\textbf{Output:}}

\noindent{$\bullet$} Forecast flow rates $\hat{\mathcal{Y}} =$MNDE$(\mathcal{X},\theta) \in \mathbb{R}^{n \times l'}$

\noindent{\textbf{Objective:}}

\noindent{$\bullet$} Optimize the parameters $\theta$ of MNDE such that the error between forecast flow rates  $\hat{\mathcal{Y}}$ and the ground truth $\mathcal{Y}$ is minimized

\section{Related Works on Traffic Forecasting}
\label{sec:related_works}

In this section, we discuss various traffic transportation forecasting approaches, encompassing techniques ranging from traffic-focused machine learning methods to advanced deep learning strategies and neural differential equation methods.

\subsection{Statistics and Machine Learning Approaches}
There are a variety of traditional statistical and machine learning methods for traffic transportation forecasting.  Some prominent example models are  (1) Autoregressive integrated moving average model (ARIMA) 
\cite{van1996combining,arima_traffic}
integrates the autoregressive model with moving average operation; (2) Seasonal autoregressive integrated moving average (SARIMA) \cite{williams2003modeling} 
adds a specific ability to ARIMA for the recognition of seasonal patterns; and (3) K-nearest neighbor model (KNN)
% cover1967nearest,
\cite{ZHANG2013653} which predict traffic of a node based on its k-nearest neighbors. A common shortcoming of these models is their focus on temporal dependencies while often neglecting spatial factors, which are crucial in understanding traffic patterns across different areas. Furthermore, their reliance on human-designed features restricts their capability to autonomously discover more predictive and task-specific features, limiting their adaptability and effectiveness.

\subsection{Deep Learning Approaches}
There are extensive works on deep learning for traffic flow forecasting. Due to space limits, we only summarize some representative works. Diffusion convolutional recurrent neural network (DCRNN) \cite{veeriah2015differential}, for instance, uses diffusion graph convolutional network (GCN) \cite{kipf2016semi}, which captures spatial correlations through bi-directional exchanges of information across nodes, with a GRU-based network \cite{chung2014empirical} adept at temporal correlation analysis. However, while GRU excels in short-term correlations, it's less effective for long-range dependencies. Addressing this, spatio-temporal graph convolutional networks (STGCN) \cite{yu2017spatio} and graph wavenet for deep spatial-temporal graph modeling (GraphWaveNet) \cite{wu2019graph} apply convolutional operations in both spatial and temporal dimensions, offering more nuanced insights into traffic patterns. Deep spatio-temporal residual networks (STResNet) \cite{zhang2017deep} takes the deep Resnet structure to capture temporal closeness, period, and trend properties of traffic patterns. Spatial-temporal synchronous graph convolutional network (STSGCN) further advances this by focusing on localized spatio-temporal subgraphs, enhancing the model's ability to understand immediate, localized correlations. However, since the model formulation does not incorporate global information, this model still had limitations when it came to long-term forecasting. In addition to the spatial graphs from predefined road networks, spatial-temporal fusion graph neural networks (STFGNN) \cite{li2020spatial} later introduce the use of dynamic time warping \cite{muller2007dynamic} for data-driven spatial networks which helped the model to learn representations from different data-driven and domain-driven views. Automated dilated spatio-temporal synchronous graph network (Auto-DSTSG) \cite{jin2022automated} implements the auto ML technique to search an optimal graph structure based on the idea of spatiotemporal synchronous graph modeling from STSGCN. Dynamic graph convolutional recurrent network (DGCRN) \cite{li2023dynamic} utilizes the generated graph, which can effectively cooperate with
a pre-defined graph while improving the prediction performance. Spatio-temporal differential equation network (STDEN) \cite{ji2022stden} borrows the idea in physics and introduces potential energy fields \cite{thermodynamics1992doe} and derives a differential equation to further describe the physical dynamics of the traffic potential energy fields. Graph multi-attention
network (GMAN) \cite{zheng2020gman} adapts an encoder-decoder architecture, where both the encoder and the decoder consist of multiple spatio-temporal
attention \cite{vaswani2017attention} blocks to model spatio-temporal pattern.  Spatio-temporal joint graph convolutional networks (STJGCN) \cite{zheng2023spatio} encompasses the construction of both
pre-defined and adaptive spatio-temporal joint graphs between time steps, which represent comprehensive and dynamic spatio-temporal correlations. Spatio-temporal meta-graph learning for traffic forecasting (MegaCRN) \cite{jiang2023spatio} utilizes the attention mechanism to build a novel Meta-Node Bank to achieve an excellent embedding for the downstream network. Spatio-temporal self-supervised learning for traffic flow prediction (ST-SSL) \cite{ji2023spatio} is benefited from self-supervised learning, by adding spatial and temporal heterogeneity-aware self-supervised signals. Spatio-temporal adaptive embedding makes vanilla transformer SOTA for traffic forecasting (STAEformer) \cite{liu2023spatio} presents the spatio-temporal adaptive embedding that can yield outstanding results with vanilla transformers. Deep expansion learning for periodic time series forecasting (DEPTS) \cite{fan2022depts} expends the periodic state capture the complicated temporal pattern. The above works are the univariate forecasting, which is to predict the future value of a single variable based solely on its historical data. Whereas, multivariate forecasting involves using historical data from multiple variables to predict the future value of a target variable. They assume the target is influenced by multiple factors and there are complex relationships between the variables. Rethinking Multivariate Time Series Forecasting from a Pure Graph Perspective (FourierGNN) \cite{yi2023fouriergnn} treats multivariate time series from a pure graph perspective and proposes the Fourier graph operator to perform matrix multiplication in Fourier space. Spatial-temporal identity: a simple yet effective baseline for multivariate time series forecasting (STID) \cite{shao2022spatial} proposes an effective baseline for multi-variant time series forecasting by attaching spatial and temporal identity information, based on simple multi-layer perceptrons.

\subsection{Neural Differential Equation Methods}

Neural Differential Equations (NDEs) \cite{DBLP:journals/corr/abs-2202-02435} can optimize neural networks in a continuous manner using differential equations. They are capable to do high-capacity function approximation, utilize continuous depth neural networks. There are a few types of NDEs: (1) neural ordinary differential equation (NODE) such as ODE-based RNN models \cite{node_rnn1,lechner2020learningy} (e.g., ODE-RNN, ODE-GRU, ODE-LSTM) and latent ODE models \cite{rubanova2019latent}; (2) neural controlled differential equation (NCDE) \cite{NEURIPS2020_4a5876b4,ncde_online} which is usually used to learn functions for the analysis of irregular time-series data; and (3) neural stochastic differential equation \cite{nsde_gan,nsde_gradient} which is usually employed for generative models that can represent complex stochastic dynamics.

More recently, Spatial-temporal graph ode networks for traffic flow forecasting (STGODE) \cite{fang2021spatial} brings us the attention to the NODE application in transportation forecasting area. It breaks through the limit of network depth and improves the capacity of extracting longer-range spatial-temporal correlations. Based on STGODE, Graph neural controlled differential equations for traffic forecasting (STGNCDE) \cite{choi2022graph} provides a new perspective of continuous depth generation by employing NCDE to capture the spatio-temporal dependencies, where the controlled term helps the model gain more powerful representations. 
\section{Proposed Approach}
\label{sec:model}

\begin{figure*}[!ht]
\centering

\includegraphics[width=1.0\textwidth]{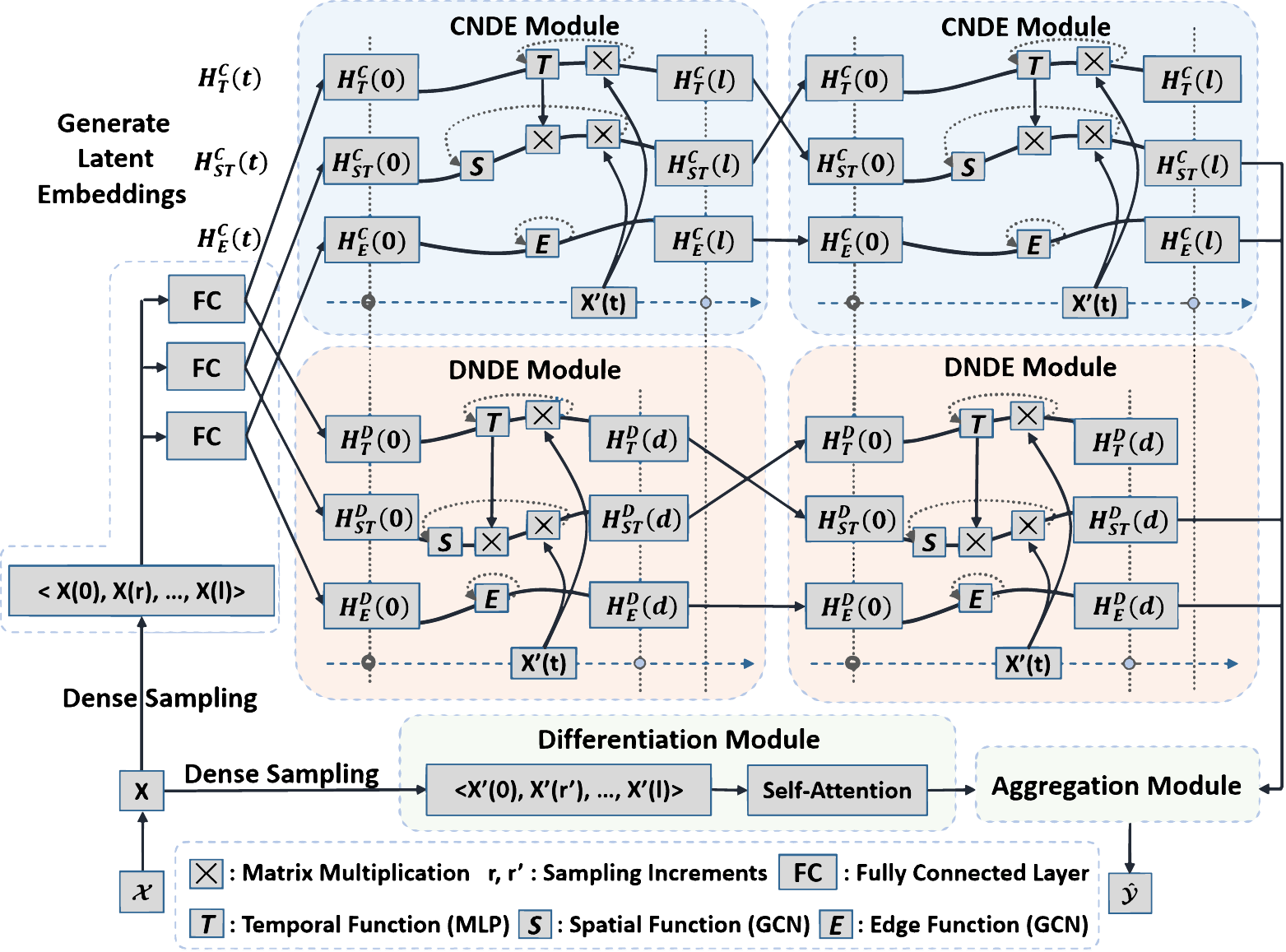}

\caption{An overview of the Multi-View Neural Differential Equation (MNDE) framework, which comprises four distinct modules. (A) Current Neural Differential Equations (CNDE) Module, (B) Delayed Neural Differential Equations (DNDE) Module, (C) Differentiation Module (DM), and (D) Aggregation Module. }
\label{fig:whole_model}
\end{figure*}

This section introduces our Multi-View Neural Differential Equation (MNDE) framework, illustrated in 
Fig. \ref{fig:whole_model}. We give an overview below before providing the details for each of its components. The framework begins by interpolating discrete flow-rate measurements into continuous time functions and then learns three latent embeddings (through fully connection layers) on top of dense samples from $X$. These embeddings capture different latent dynamic patterns of the traffic flows, including the temporal dynamics ($H_T^C(t)$), spatio-temporal dynamics ($H_{ST}^C(t)$), and dynamic edge interactions ($H_E^C(t)$), respectively, where superscript $C$ stands for {\it current pattern}. Each of these three current dynamic patterns is modeled by an NDE (shown in the top of the figure). The spatio-temporal dynamics are computed based on the temporal dynamics ($T$ in the figure) and the spatial dynamics from a graph convolutional network component ($S$ in the figure). Another graph convolutional network component ($E$ in the figure) captures the edge dynamics. We integrate these components together to construct the current neural differential equations module (CNDE). We feed the output of CNDE back as the input for a number of times --- 3 times in our experiments but 2 times in the figure to save space --- to create deeper networks for better capturing of the dynamic patterns. To capture the delayed propagation effect in traffic flow dynamics, we introduce a new delayed neural differential equations module (DNDE), which extends the CNDE module with a delayed factor in temporal integration to capture the delayed temporal dynamics ($H_T^D(t)$), delayed spatio-temporal dynamics ($H_{ST}^D(t)$), and delayed edge interactions ($H_E^D(t)$), where superscript $D$ stands for {\it delayed pattern}. Again, we feed the output of DNDE back as the input a number of times --- 3 times in our experiments but 2 times in the figure to save space --- for deeper networks. We also add a differentiation module (DM), which uses self-attention on temporal gradients to capture the abrupt shift patterns. The learned embeddings of different latent dynamics are finally aggregated to make the final forecasting.

\subsection{Temporal Interpolation and Latent Embeddings}
After interpolating input measurements from discrete times to continuous time through the cubic spline method, we perform a dense temporal sampling for a much expanded input than the original measurements, as is illustrated in the figure with the sampling steps $r, r' < 1$; we set $r$ and $r'$ to 1/3 in our experiments. The dense samples are fed into three separate fully connection layers to generate embeddings, $H_T^C(0)\in \mathbb{R}^{n \times c}$, $H_{ST}^C(0)\in \mathbb{R}^{n \times c}$, and $H_E^C(0)\in \mathbb{R}^{n \times n \times c'}$, where $c$ and $c'$ are the embedding dimensions. These embeddings become the initial states of three NDEs for temporal dynamics $H_T^C(t)$, spatio-temporal dynamics $H_{ST}^C(t)$, and dynamic-edge patterns $H_E^C(t)$, respectively. The dimension of $H_E^C(0)$, i.e., $n \times n \times c'$, is different from the other two due to the need to capture {\it pair-wise} location interactions. This process is first done for the Current NDE module and then repeated for the Delayed NDE module to produce latent embeddings of $H_T^D(0)$, $H_{ST}^D(0)$, and $H_E^D(0)$ as the initial values of three delayed NDEs, $H_T^D(t)$, $H_{ST}^D(t)$, $H_E^D(t)$, respectively.

\subsection{Current Neural Differential Equations (CNDE) Module}
The CNDE module consists of three NDEs for temporal dynamics, spatio-temporal dynamics, and dynamic-edge patterns, respectively, which are explained below.

{\bf Temporal dynamics:} The temporal function $T$ is a crucial component in the recursive implementation of the NDE for the latent temporal dynamics $H_T^C(t)$, as in (\ref{eq:t}), to generate the final result $H_T^C(l)$. $T$ takes $H_T^C(t)$, initially $H_T^C(0)$, as input and performs three fully connected layers with a skip connection to produce a tensor of dimension $\mathbb{R}^{n \times c}$. We do not use a recurrent neural network layer in $T$, as the temporal information is incorporated by the multiplier $X'(t)$ in the NDE model.

\begin{align}\label{eq:t}
H_T^C(l)&=H_T^C(0)+\int_{0}^{l}T(H_T^C(t))X'(t)\mathrm{d} t 
\end{align}

{\bf Spatial-Temporal dynamics:} To capture the spatio-temporal dynamics, we multiply the spatial function $S$ and temporal function $T$ within the second NDE. Following \cite{choi2022graph}, $S$ is designed based on a graph neural network (GCN) structure \cite{defferrard2016convolutional}. The final results of spatial-temporal dynamics, $H_{ST}^C(l)$, is computed by (\ref{eq:st}), following the general NDE in (\ref{NCDE:1}), with $f_c$ built by multiplying $S$ and $T$. As shown in (\ref{eq:st}), $S$ as a function of $H_{ST}^C(t)$ consists of a GCN layer and a fully connected layer ($\mathcal{F}$), followed by a skip connection. The adjacency matrix $A_S$ and the parameters $W_S$, $b_{S}$ are learnable. Here $\sigma$ is a non-linear activation function (e.g., ReLU). 

\begin{align}\label{eq:st}
H_{ST}^C(l)&=H_{ST}^C(0)+\int_{0}^{l}S(H_{ST}^C(t))T(H_{ST}^C(t))X'(t)\mathrm{d} t \\
S(H_{ST}^C(t))&=\mathcal{F}(GCN(H_{ST}^C(t)))+\mathcal{F}(H_{ST}^C(t))\nonumber\\
&=\mathcal{F}(\sigma(A_{S}H_{ST}^C(t)W_{S}+b_{S}))+\mathcal{F}(H_{ST}^C(t))
\end{align}

{\bf Edge dynamics:}

Traffic flows at different locations are not isolated in a real-world road system, where the dynamic patterns of some flows are associated under similar temporal conditions (such as peak hours), similar residential/commercial environment, or same vehicles that travel through them. We model such potential association with an artificial edge between each pair of locations and capture the {\it dynamic edge patterns} with one extra NDE, $H_E^C(t)$. 
As shown in (\ref{eq:E-CNDE}), we formulate an edge function $E$ in a similar construction as the $S$ function, consisting of a GCN layer and a fully connected layer ($\mathcal{F}$), followed by a skip connection.  The difference is that input latent embedding, $H_E^C(t)$ with initial value of $H_E^C(0)$, is of a different size, $n \times n \times c'$, in order to reflect pair-wise features between locations. 
Our method is different from the existing work \cite{yu2017spatio,li2018diffusion,wu2019graph,li2020spatial} in that we use dynamic edge features in the GCN operation, instead of 
considering edges in a static or fixed adjacency matrix. The advantage is that such a design can learn the edge-level temporal knowledge from the node (location) level representation.

\begin{align}
\label{eq:E-CNDE}
H_E^C(l)&=H_E^C(0)+\int_{0}^{l}E(H_E^C(t))\mathrm{d} t \\
\label{eq:E-DNDE}
E(H_E^C(t))&=\mathcal{F}(GCN(H_E^C(t)))+\mathcal{F}(H_E^C(t)) \nonumber\\
&=\mathcal{F}(\sigma(A_{E}H_E^C(t)W_{E}+b_{E}))+\mathcal{F}(H_E^C(t))
\end{align}

\subsection{Delayed Neural Differential Equations Module}
In order to characterize the delayed propagation of traffic flow dynamics (where traffic from an upstream location affects a downstream location at a later time), we introduce a new delayed NDE module. It is known to the transportation community \cite{sipahi2010deterministic,horowitz2002internodal} that traffic flow dynamics are determined by both the current state and the delayed state in a road system. Consider an example where congestion is developed at one segment of a highway. As the vehicles slow down, those behind will slow down. This chain reaction of slowing down propagates backward along the highway. Consequently, drivers further downstream will experience a delayed slowdown (i.e., a smaller flow rate) and even congestion. Therefore, the downstream traffic dynamics are determined by both the delayed state from upstream and the current local state. None of the prior work in deep learning based traffic forcasting \cite{grigsby2021long,song2020spatial,choi2022graph} has considered the delayed state. 

The Delayed NDE (DNDE) module is illustrated under the current NDE model (CNDE) in Figure \ref{fig:whole_model}.  The design of DNDE, as shown in (\ref{eq:dt})-(\ref{eq:de}), is controlled by $X'(t)$ with $t \in [0, d]$, where $d$ should be set less than $l$, allowing the integration to focus on the data into the past, without the interference of the data closer to the present, i.e., in the range of $t \in (d, l]$. Through independent validation datasets, we find that  $d<l/3$ gives adequate discrepancy between the patterns captured by DNDE and CNDE.

\begin{align}\label{eq:dt}
H_T^D(d)&=H_T^D(0)+\int_{0}^{d}T(H_T^D(t))X'(t)\mathrm{d} t \\
H_{ST}^D(d)&=H_{ST}^D(0)+\int_{0}^{d}S(H_{ST}^D(t))T(H_{ST}^D(t))X'(t)\mathrm{d} t \\
\label{eq:de} H_E^D(d)&=H_E^D(0)+\int_{0}^{d}E(H_E^D(t))\mathrm{d} t 
\end{align}

\subsection{Differentiation Module}

In order to capture abrupt shift patterns like sudden road closure due to car accidents or severe weather in a road system (unforeseen incidents that unfold suddenly), we design a differentiation module to learn from the local temporal gradients of traffic flow data. The module aims to conduct fine-grained analysis to capture nuanced fluctuations and interactions. This is different from the existing work that focuses on regularly occurring events such as rush hour congestion \cite{wu2019graph,Jiang_Wang_Yong_Jeph_Chen_Kobayashi_Song_Fukushima_Suzumura_2023}. Specifically, we concatenate the sampled temporal gradients of traffic flow data, denoted as $D = [X'(0); X'(r'); ...; X'(l)] \in \mathbb{R}^{n \times \frac{l}{r'}}$, where $r'$ (e.g., 1/3) is the increment step for sampling. Then we apply a self-attention layer \cite{vaswani2017attention}, as shown in (\ref{eq:AM1}), on the input $D$ to produce an output $D'$, $ \mathbb{R}^{n \times l'}$ given by (\ref{eq:AM2}). Recall that $l'$ is the length of the output for each flow in the problem statement. The self-attention layer focuses its learning on the nuances, where the weight matrices $W_q \in \mathbb{R}^{\frac{l}{r'} \times c'}$, $W_k \in \mathbb{R}^{\frac{l}{r'} \times c'}$, and $W_v \in \mathbb{R}^{\frac{l}{r'} \times l'}$ represent the attention operations for query $Q$, key $K$, and value $V$, respectively. The number of attention heads is set by $h$, and $c'$ is the embedding dimension.

\begin{equation}
\begin{aligned}
\label{eq:AM1}
Q= DW_q+b_q, \,\,
K= DW_k+b_k, \,\,
V= D W_v+b_v 
\end{aligned}
\end{equation}

\begin{equation}
\label{eq:AM2}
D'=softmax\,(\,\sqrt{\frac{c'}{h}}\,(\,Q^T K\,)) \,\,V
\end{equation}

\subsection{Aggregation Module}

We aggregate the final latent embeddings from the CNDE module, the DNDE module, and the differentiation module to make our long-term flow rate forecast. We first use CNN and MLP to transform the outputs of the CNDE and DNDE Modules, i.e., $H_{ST}^C(l)$, $H_{E}^C(l)$, $H_{ST}^D(d)$, and $H_{E}^D(d)$, to the space of $\mathbb{R}^{n \times l'}$. They are denoted as $p_0$, $p_1$, $p_2$ and $p_3$, respectively, after transformation. We do not use $H_{T}^C(l)$ and $H_{T}^D(d)$ because the temporal dynamic patterns are already incorporated into $H_{ST}^C(l)$ and $H_{ST}^D(l)$. Denote $D'$ as $p_4$. The aggregation formula is given in (\ref{eq:aggregation}), where each transformed output is multiplied by the sum of other transformed outputs that are first normalized using the softmax operation. This aggregation technique offers several advantages, including the ability to choose critical forecasting features and non-linear aggregation.
This non-linear operation \cite{liu2023graphbased} is used to account for correlations in the higher dimensions.

\begin{table*}[!t]
    \centering
    \caption{Performance comparison. The \textbf{best} results are in bold and the  \underline{second-best} are underlined. }
    \label{tab:results}
    \resizebox{1.0\textwidth}{!}{%
    \begin{tabular}{|c|c|c|c|c|c|c|c|c|c|c|c|c|c|c|c|c|c|c|}
        \hline
        \multirow{3}{*}{Model} & \multicolumn{9}{c|}{PEMS04} & \multicolumn{9}{c|}{PEMS08} \\ \cline{2-19} & \multicolumn{3}{c|}{2nd hour} & \multicolumn{3}{c|}{4th hour} & \multicolumn{3}{c|}{8th hour}& \multicolumn{3}{c|}{2nd hour} & \multicolumn{3}{c|}{4th hour} & \multicolumn{3}{c|}{8th hour}\\ \cline{2-19}
        & MAE & RMSE & MAPE & MAE & RMSE & MAPE & MAE & RMSE & MAPE & MAE & RMSE & MAPE & MAE & RMSE & MAPE & MAE & RMSE & MAPE\\ \hline
        GraphWaveNet \cite{wu2019graph} & 50.37 & 73.82 & 40.71 & 81.34 & 113.85 & 86.65 & 106.59 & 143.33 & 151.47 & 38.04 & 57.86 & 28.62 & 62.45 & 90.9 & 64.68 & 83.70 & 114.66 & 137.39\\ \hline
        STGCN \cite{yu2017spatio} & 33.33 & 48.46 & 20.94 & 48.02 & 83.56 & 28.62 & 37.35 & 55.38 & 30.92 & 26.24 & 38.25 & 15.37 & 35.46 & 48.88 & 21.35 & 48.93 & 68.77 & 35.97\\ \hline
        STSGCN \cite{song2020spatial} & 29.59 & 46.31 & 19.55 & 34.61 & 53.82 & 24.58 & 38.16 & 58.49 & 31.61 & 23.25 & 36.21 & 15.21 & 26.29 & 41.55 & 17.19 & 33.29 & 51.52 & 23.74\\ \hline
        STFGNN \cite{li2020spatial} & 25.16 & 40.21 & 19.18 & 31.73 & 49.51 & 26.73 & 37.31 & 57.72 & 31.06 & 23.27 & 36.78 & 14.97 & 27.12 & 42.75 & 19.12 & 34.62 & 53.82 & 25.01\\ \hline
        STGODE \cite{fang2021spatial} & 27.84 & 41.76 & 20.58 & 33.26 & 49.49 & 25.76 & 36.83 & 54.03 & 33.77 & 22.94 & 35.53 & 15.01 & 29.20 & 44.62 & 20.18 & 36.36 & 54.74 & 29.23\\ \hline
        MegaCRN \cite{Jiang_Wang_Yong_Jeph_Chen_Kobayashi_Song_Fukushima_Suzumura_2023} & 24.08 & 37.95 & 17.43 & 28.74 & 45.20 & 20.50 & 35.10 & 55.26 & 23.51 & 20.57 & 32.19 & 14.37 & 25.52 & 39.89 & 17.14 & 33.40 & 51.19 & 22.39\\ \hline
        Z-GCNETs \cite{chen2021z}& 23.53 & 37.84 & 15.40 & 26.69 & 42.57 & \underline{18.18} & 32.07 & 49.84 & 23.55 & 21.32 & 33.52 & \underline{13.55} & \underline{23.35} & \underline{35.91} & \underline{16.02} & 28.78 & 43.51 & 21.97\\ \hline
        FourierGNN \cite{yi2024fouriergnn} & 41.04 & 60.00 & 32.30 & 54.11 & 77.50 & 43.33 & 69.43 & 95.04 & 71.44 & 35.44 & 49.58 & 33.25 & 46.04 & 63.75 & 45.05 & 59.28 & 80.67 & 53.81 \\ \hline
        ST-SSL \cite{ji2023spatio}&  25.81 & 40.19 & 20.93 & 28.69 & 45.21 & 21.67 & 32.73 & 50.64 & 24.95 & 23.42 & 34.93 & 18.60 & 26.47 & 39.52 & 20.13 & 30.69 & 45.97 & 23.28\\ \hline
        Spacetimeformer \cite{grigsby2021long} & 44.63 & 126.11 & 113.98 & 41.71 & 92.91 & 76.97 & 47.90 & 74.05 & 45.56 & 23.71 & 35.53 & 14.69 & 26.18 & 38.88 & 16.57 & 27.04 & 41.87 & 19.75  \\ \hline
        STGNCDE \cite{choi2022graph} & \underline{23.06} & \underline{36.95} & \underline{15.27} & \underline{26.44} & \underline{41.74} & 18.29 & \underline{30.27} & \underline{47.78} & \underline{22.64} & \underline{20.09} & \underline{31.39} & 13.71 & 23.99 & 36.78 & 16.22 & \underline{27.22} & \underline{41.93} & \underline{20.28} \\ \hline
        MNDE & \textbf{22.15} & \textbf{35.65} & \textbf{14.26} & \textbf{25.20} & \textbf{40.38} & \textbf{17.15} & \textbf{26.05} & \textbf{41.20} & \textbf{18.25} & \textbf{19.71} & \textbf{31.06} & \textbf{13.14} & \textbf{21.54} & \textbf{33.85} & \textbf{14.29} & \textbf{25.18} & \textbf{39.10} & \textbf{17.64}\\ \hline
    \end{tabular}
    }

\centering

\end{table*}

\begin{table}[!t]
\resizebox{1.0\columnwidth}{!}{%
\begin{tabular}{|c|c|c|c|c|c|c|c|c|c|}
\hline
\multirow{3}*{Model}  & \multicolumn{9}{c|}{PEMS03}\\ \cline{2-10}
& \multicolumn{3}{c|}{2 hour} & \multicolumn{3}{c|}{4 hour} & \multicolumn{3}{c|}{8 hour} \\ \cline{2-10} 
 & MAE & RMSE & MAPE & MAE & RMSE & MAPE & MAE & RMSE & MAPE \\ \hline
Graph WaveNet \cite{wu2019graph}& 41.34 & 62.88 & 47.44 & 67.87 & 99.09 & 94.56 & 92.96 & 126.41 & 144.32 \\ \hline
STGCN \cite{yu2017spatio} & 30.20 & 46.56 & 32.89 & 44.11 & 74.99 & 48.21 & 41.55 & 65.84 & 57.01 \\ \hline
STSGCN \cite{song2020spatial}& 27.54 & 45.48 & 31.55 & 34.36 & 57.91 & 43.37 & 39.62 & 65.35 & 55.63 \\ \hline
STFGNN \cite{li2020spatial}& 27.21 & 45.84 & 28.06 & 24.48 & 58.35 & 38.06 & 40.54 & 66.71 & 58.19 \\ \hline
STGODE \cite{fang2021spatial}& 27.36 & 42.98 & 27.90 & 35.35 & 55.88 & 33.86 & 42.40 & 66.12 & 57.97 \\ \hline
FourierGNN \cite{yi2024fouriergnn} & 34.77 & 52.22 & 45.91 & 47.38 & 69.48 & 68.83 & 60.82 & 83.61 & 102.04 \\ \hline
ST-SSL \cite{ji2023spatio} & 23.28 & 37.27 & 25.34 & 26.19 & 42.42 & 26.73 & 33.59 & 55.28 & 41.83 \\ \hline
MegaCRN \cite{Jiang_Wang_Yong_Jeph_Chen_Kobayashi_Song_Fukushima_Suzumura_2023} & \underline{19.99} & \underline{33.97} & \underline{18.53} & \underline{24.06} & \underline{41.47} & \underline{24.11} & 31.54 & 54.65 & 45.01\\ \hline
STGNCDE \cite{choi2022graph}& 22.25 & 36.78 & 21.81 & 25.86 & 43.22 & 26.17 & \underline{30.36} & \underline{50.46} & \underline{35.34} \\ \hline
MNDE & \textbf{17.91} & \textbf{30.71} & \textbf{17.40} & \textbf{23.83} & \textbf{39.86} & \textbf{22.14} & \textbf{27.37} & \textbf{47.04} & \textbf{27.57} \\ \hline

\end{tabular}
}
\end{table}

\begin{align}
\label{eq:aggregation}
\hat{\mathcal{Y}}=&\frac{1}{K(K-1)}\sum\limits_{m}^{K} {p_m} \odot (\sum\limits_{n\neq m}^{K} softmax(p_n))
\end{align}

\subsection{Loss Function}
The Huber loss function is commonly employed in regression problem scenarios, such as traffic flow forecasting. This function is a hybrid of the $L1$ and $L2$ loss functions and is well documented in literature \cite{schmidt2007fast}. The Huber loss function is composed of two parts, which form a piece-wise function: (1) robustness: a squared term that has a smooth and small gradient when the disparity between the true and predicted values is small (i.e., less than a $\delta$ threshold value), and (2) accuracy: a restricted gradient term when the true and predicted values are considerably different. We choose Huber loss due to its balance between accuracy and robustness.  The standard form of the Huber loss function is represented in equation (\ref{eq:huberloss}).

\vspace{-0.5em}
\begin{equation}
\label{eq:huberloss}
L(\hat{\mathcal{Y}},\mathcal{Y}) = \left\{  
\begin{aligned} 
&\frac{1}{2} (\hat{\mathcal{Y}}-\mathcal{Y})^2 , \,\,\,\,\,\,\,\,\,\,\,\,|\hat{\mathcal{Y}}-\mathcal{Y}| \leq \delta \\
&\delta|\hat{\mathcal{Y}}-\mathcal{Y}|-\frac{1}{2}\delta^2 , \,\, otherwise
\end{aligned}\right.
\end{equation}
where $\delta$ is a hyperparameter value set for the intended threshold; $\mathcal{Y}$ is the true future spatio-temporal data; and $\hat{\mathcal{Y}}$ is the predicted future data.

\section{Experiments and Numerical Results}\label{sec:experiments}

\subsection{Settings and Datasets} \label{subsec:settings}

\begin{table}[t]
\begin{center}
\captionsetup{font=small}
\caption{\label{tab:Data_Statistics} Basic statistics of the datasets used in our experiments.}
 \aboverulesep=0ex
 \belowrulesep=0ex
\setlength{\tabcolsep}{3pt}
\renewcommand{\arraystretch}{1.2}
\large
\centering
\rmfamily
\resizebox{\linewidth}{!}{
\begin{tabular}{ c | c  | c | c  | c | c} 
\toprule
\textbf{Data} & \textbf{PEMS03} & \textbf{PEMS04}  & \textbf{PEMS08} & \textbf{PEMS-BAY} & \textbf{METR-LA} \\
\midrule
Area & \multicolumn{5}{c}{CA, USA}\\ 
\midrule
Time Span & 9/1/2018 - & 1/1/2018 -   & 7/1/2016 - & 1/1/2017 - & 3/1/2012 -\\ 
& 11/30/2018 & 2/28/2018  & 8/31/2016 & 5/31/2017 & 6/30/2012 \\
\midrule
Time Interval & \multicolumn{5}{c}{5 min}\\ 
\midrule
Number of Locations & 358 & 307   & 170 & 325 & 207\\ 
\midrule
Number of Time Intervals & 26,208 & 16,992   & 17,856 & 52,116 & 34,272\\ 
\bottomrule
\end{tabular}}
\end{center}
\end{table}

\noindent\textbf{Dataset:} We use three widely used public traffic flow datasets, PEMS03, PEMS04, PEMS08, PEMS-BAY, and METR-LA \footnote{The datasets are provided in the STSGCN github repository at \url{https://github.com/Davidham3/STSGCN/}, and DCRNN github repository \url{https://github.com/liyaguang/DCRNN}.} which are collected from the Freeway Performance Measurement System in California, USA. The details of data statistics are shown in Table. \ref{tab:Data_Statistics}. Both datasets measure the flow rate data, i.e., number of passing vehicles at each location during each 5-min interval. Following the prior work \cite{song2020spatial,choi2022graph}, we pre-process the datasets by the z-score normalization before using them as input. The z-score normalization subtracts the mean rate from each flow rate in the dataset and then divides the result by the rates' standard deviation. We split each dataset into three subsets in 3:1:1 ratio for training, validation, and testing. We use the past 1 hour of flow data to predict the future 2, 4, and 8 hours of flow data. Because the time interval for flow-rate measurement is five minutes, each hour has 12 flow-rate data points. 

Experiment were conducted on server with Intel(R) Xeon(R) Gold 6142 CPU @ 2.60GHz, 280 GB of main memory, and NVIDIA A100 GPU with 80 GB memory.

\noindent \textbf{Hyperparameters:}  For dataset PEMS03,  $n = 358$; for dataset PEMS04,  $n = 307$; for dataset PEMS08, $n=170$; for dataset PEMS-BAY, $n=325$; for dataset METR-LA, $n=207$. Increment steps, $r$ and $r'$, are set to $1/3$. For embedding dimension, $c = 64$, and $c' = 32$. In the DNDE module, $d = 2$. In the Differentiation Module, $h = 4$. We use Huber Loss \cite{schmidt2007fast} as the loss function. Both the CNDE module and the DNDE module are looped three times, with their outputs fed back as the inputs of the modules. Hence, $l=12$ and $l'=96$.  During training, we use the learning rate at $10^{-3}$ and the weight decay at $10^{-3}$ for both datasets. The optimizer is AdamW \cite{loshchilov2018decoupled}. Batch size is 64 for every dataset. To solve this NDE problem, we use torchdiffeq \cite{chen2018neuralode} with runge-kutta 4-th order method \cite{butcher1987numerical}.  The relative tolerance and absolute tolerance for NDE solver are $10^{-9}$ and $10^{-7}$. We use Mean Absolute Error (MAE), Root Mean Squared Error (RMSE), and Mean Absolute Percentage Error (MAPE) as the performance metrics to evaluate the performance of MNDE. For the other baselines, we use the recommended hyper-parameters provided in their paper and code.  

Given the ground truth $\mathcal{Y}= [\mathcal{Y}_{i,j}, i \in \mathcal{N}, j \in \{0,1,..,l'-1\} ]^T$ and the forecast $\hat{\mathcal{Y}}= [\hat{\mathcal{Y}}_{i,j}, i \in \mathcal{N}, j \in \{0,1,..,l'-1\} ]^T$, 

\begin{align}
\label{eq:mae}
\text{MAE at time\ } j = & \frac{1}{n}\sum\limits_{i=1}^{n}\left|\mathcal{Y}_{i,j} -\hat{\mathcal{Y}}_{i,j}\right| \\
\text{MAPE at time\ } j =& \Big(\frac{1}{n}\sum\limits_{i=1}^{n}\left|\frac{\mathcal{Y}_{i,j} -\hat{\mathcal{Y}}_{i,j}}{\mathcal{Y}_{i,j}}\right| \Big) * 100\%\\
\text{RMSE at time\ } j = & \sqrt{\frac{1}{n}\sum\limits_{i=1}^{n}\Big(\mathcal{Y}_{i,j} -\hat{\mathcal{Y}}_{i,j}\Big)^2}
\end{align}

\subsection{Prior Work}

For all prior works, we run their codes and follow their recommended configuration. In order to fit in the long-term forecasting scenario, we set all the under models with history length $l=12$, and future length $l'=96$. All results are obtained by averaging the outcomes of four separate experiments.

\begin{itemize}

\item{\textbf{GraphWaveNet} \cite{wu2019graph}: Graph WaveNet integrates adaptive graph convolution with 1D dilated casual convolution to capture spatio-temporal dependencies.}

\item{\textbf{STGCN} \cite{yu2017spatio}: Spatio-Temporal Graph Convolutional Network combines graph structure convolutions with 1D temporal convolutional kernels to capture spatial dependencies and temporal correlations, respectively.}

\item{\textbf{STSGCN} \cite{song2020spatial}: Spatio-Temporal Synchronous Graph Convolutional Networks decompose the problem into multiple localized spatio-temporal subgraphs, assisting the network in better capturing of spatio-temporal local correlations and consideration of various heterogeneities in spatio-temporal data. }
\item{\textbf{STFGNN} \cite{li2020spatial}: Spatio-Temporal Fusion Graph Neural Networks uses Dynamic Time Warping (DTW) algorithm to gain features, and follow STSGCN \cite{song2020spatial} in using sliding window to capture spatial, temporal, and spatio-temporal dependencies.}

\item{\textbf{STGODE} \cite{fang2021spatial}: Spatio-Temporal Graph ODE Networks attempt to bridge continuous differential equations to the node representations of road networks in the area of traffic forecasting.}

\item{\textbf{MegaCRN} \cite{jiang2023spatio}: Meta-Graph Convolutional Recurrent Network utilize the attention mechanism to build a novel Meta-Node Bank to achieve an excellent embedding for downstream network.}

\item{\textbf{Z-GCNETs} \cite{chen2021z}: Time Zigzags at Graph Convolutional Networks attempt to bridge continuous differential equations to the node representations of road networks in the area of traffic forecasting.}

\item{\textbf{Spacetimeformer} \cite{grigsby2021longrange}: Long-Range Transformers tokenize the spatio-temporal series and then learn interactions between space, time, and value information jointly along this extended sequence. }

\item{\textbf{FourierGNN}\cite{yi2023fouriergnn}: This model rethinks time series forecasting from a pure graph perspective and proposes the Fourier graph operator to perform matrix multiplication in Fourier space.}

\item{\textbf{ST-SSL}\cite{ji2023spatio}: The Spatio-Temporal Self-Supervised Learning approach enhances traffic pattern representations with auxiliary self-supervised learning paradigms.}

\item{\textbf{STGNCDE} \cite{choi2022graph}: Spatio-Temporal Graph NCDE Networks first introduce NCDE to traffic forecasting area. It utilizes the power of controlled path and its characteristic of differentiable to achieve a excellent performance. }
\end{itemize}

\subsection{Model Performance}
% \vspace{-0.5em}
 
% answers \textbf{RQ1} and 
Table 1 presents the comparison between the proposed MNDE and the prior work in terms of MAE, MAPE and RMSE at the end of the 2nd hour, the 4th hour and the 8th hour (after the input flow-rate measurements).

Our proposed \modelname demonstrates superior performance across two datasets, three forecasting horizons, and three evaluation metrics. For example, at the 8th-hour time interval, our model exhibits significant enhancements in comparison to the second-best model, with $24\%$ and $15\%$ improvement observed in the MAPE metric.

Basic GCN models (GraphWaveNet and STGCN) perform poorly in long-term forecasting. This is likely due to the constraints imposed by limited network depth and the associated risk of over-smoothing \cite{graph-over-smoothing1,graph-over-smoothing2}.

Dynamic-aware GCNs (STSGCN and STFGNN) perform somehow better than basic GCNs. This can be explained by the fact that STSGCN captures localized temporal dynamics within distinct time periods and constructs an advanced graph based on these temporal segments. On top of STSGCN, STFGNN models temporal similarities among locations and integrates another dynamic time warping graph \cite{berndt1994using}.

The existing spatio-temporal long-term forecasting method, Spacetimeformer, yields inconsistent results in the context of PEMS04. This instability may stem from unpredictable events, to be explained with Figure \ref{fig:node_vis} (Location 7).

MegaCRN has an outstanding graph modeling structure but its design is aligned for short-term forecasting. Z-GCNETs takes advantage of time-aware persistent homology \cite{DBLP:journals/corr/abs-0812-0197}.
STGNCDE is the first to apply (\ref{NCDE:1}) in traffic forecasting, incorporating a controlled term $X'(t)$ and infusing the model with heightened temporal dynamics. It generally performs better than other related work due to its use of a neural differential equation to capture temporal dynamics. However, its performance is worse than our MNDE because the latter captures several additional latent dynamic patterns, including delayed propagation, more spatial dynamics, and local trend, as the ablation study will demonstrate next.

% \subsection{RQ2: Ablation Study}
\subsection{Ablation Study}\label{subsec:ablation}

\begin{figure}[!t]
\centering
\includegraphics[width=\columnwidth]{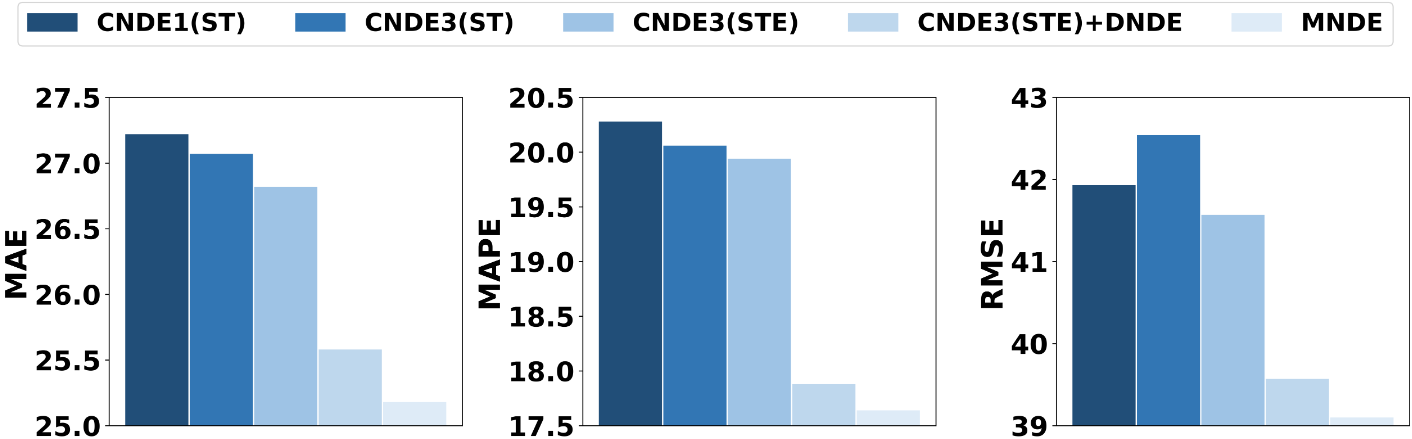}

\centering\caption{Ablation study, 8th hour time interval, PEMS08 dataset}

\label{fig:ablation_study}
\end{figure}
To investigate the effect of different components of \modelname, we conduct ablation experiments on PEMS08 with several different variants of our model.
\textbf{CNDE1(ST):} We interpolate the discrete flow data into continuous flow data and fed it into the model which only includes one pass of a simplified CNDE Module with only the temporal dynamics ($T$) and the spatial-temporal dynamics ($ST$). 
\textbf{CNDE3(ST):} It is the same s CNDE1(ST) except that the CNDE module is looped three times. \textbf{CNDE3(STE):} It uses the complete version of CNDE with the edge dynamics ($E$). The output of the edge and spatio-temporal dynamics are aggregated by the aggregation module. \textbf{CNDE3(STE)+DNDE:} It uses both CNDE module and the DNDE module, whose outputs are aggregated. \textbf{\modelname:} It adds differentiation module to capture the localized abrupt shift patterns. 

Fig. \ref{fig:ablation_study} presents the values of MAE, MAPE and RMSE under the variants of \modelname over PEMS08 dataset at the 8th hour time interval. The results show that the DNDE module and the differentiation module contribute the most in reducing the forecast errors, as \modelname and CNDE3(STE)+DNDE improve accuracy significantly over their respective predecessors, whereas the edge dynamics and the looping of the CNDE module still contribute, though by smaller amounts, as CNDE3(STE) and CNDE(STE) improve accuracy incrementally over their respective predecessors. This highlights the importance of capturing the delayed propagation of traffic conditions and the local traffic trends, which we are the first to introduce in long-term traffic forecasting. 

\begin{figure}[!h]

\centering
\includegraphics[width=0.7\columnwidth]{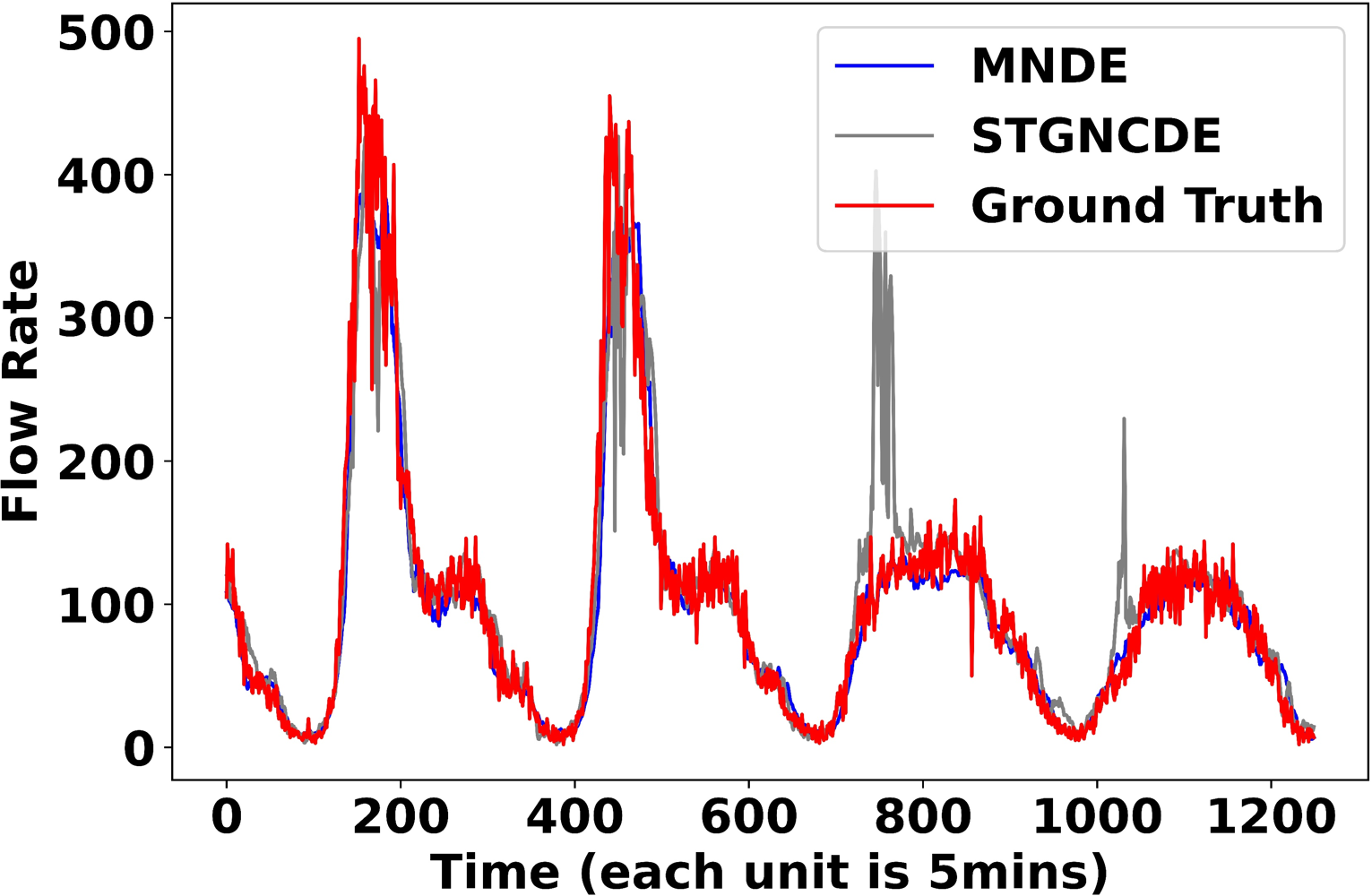}
\caption{ Comparison of traffic flow forecasting between our proposed \modelname and STGNCDE on PEMS04 dataset.
}
\label{fig:node_vis}
\end{figure}

% \textit{Case Study}:
\subsection{MNDE v.s. STGNCDE}
%\input{tables_figs_tex/embedding.tex}

% \vspace{-0.5em}
We conduct a case study that compares the proposed MNDE and the best existing work STGNCDE at a randomly chosen location from the PEMS04 road network. The location has 4 days of flow rate measurements. Figure 3 shows the long-term flow rate prediction of MNDE (STGNCDE) over time at the location, together with the ground truth, where each prediction is produced by using one hour input data that is of 8 hours ago, for long-term forecasting.  We can see that the grey line (flow-rate predictions by STGNCDE) sometimes deviates significantly from the red line of ground truth, while the blue line (predictions by MNDE) follows the ground truth much better. The reason is that the design of STGNCDE only contains CNDE (without the explicit edge dynamics $E$), while \modelname contains CNDE, DNDE and differentiation module (DM), with DNDE capturing delayed propagation of traffic dynamics and DM capturing localized traffic trend.

\subsection{Robustness Study}

\begin{table}[!t]

\caption{ Robustness comparison of the models used in ablation study on PEMS08 dataset. }
\label{tb:missing}
\resizebox{0.9\columnwidth}{!}{%
\centering
\begin{tabular}{ccccc}

\hline Model & Missing Rate & MAE & RMSE & MAPE \\
\hline CNDE(ST) & & 28.45 & 44.01 & 20.86  \\
CNDE3(ST) & \multirow{3}{*}{$10 \%$} & 27.23 & 43.88 & 20.55  \\
CNDE3(STE) & & 27.65 & 43.16 & 19.72  \\
CNDE3(STE)+DNDE & & 26.34 & 41.76 & 18.69  \\
MNDE & & 26.02 & 40.53 & 18.05  \\
\hline CNDE(ST) & & 29.73 & 44.88 & 21.57  \\
CNDE3(ST) & \multirow{3}{*}{$30 \%$} & 28.14 & 43.92 & 20.76  \\
CNDE3(STE) & & 27.75 & 43.24 & 20.01 \\
CNDE3(STE)+DNDE & & 26.70 & 41.78 & 18.59 \\
MNDE & & 26.57 & 41.35 & 18.97  \\
\hline CNDE(ST) & & 29.82 & 46.14 & 20.86 \\
CNDE3(ST) & \multirow{3}{*}{$50 \%$} & 28.29 & 43.97 & 20.19 \\
CNDE3(STE) & & 27.92 & 43.81 & 19.68  \\
CNDE3(STE)+DNDE & & 27.29 & 42.97 & 19.19  \\
MNDE & & 26.95 & 41.86 & 18.58  \\
\hline

\end{tabular}
}
\centering
\end{table}

To evaluate the robustness of our MNDE, we perform an experiment where we partially remove some of the input data by randomly substituting some original flow rate measurements  with ``NaN", simulating malfunctioned sensors with missing measurements or rejected measurements due to too much noise.  Following \cite{kidger2020neural}, we handle ``NaN"s by replacing them with interpolated values between valid neighboring measurements based on the cubic spline method.

To explore the impact of different missing rates, we vary the percentage of flow data that are missing from $10\%$ to $50\%$. We adopt the variants of our model from the ablation study to investigate the robustness of different components in MNDE. Table 2 presents the performance of the variants in terms of MAE, RSME and MAPE at the 8th hour time interval under different missing rates, $10\%$, $30\%$ and $50\%$, on PEMS08 dataset. The baseline results with zero missing rate can be found in Table 1. The table shows that MNDE and its components exhibit strong robustness because the performance of all variants degrades modestly with increased missing rates, even to 50\%. Additionally, MNDE can still beat the prior works even facing the largest missing rate (e.g. $50\%$).

\subsection{Different Length of Input Data}\label{sec:length}

\begin{figure}[!t]
\centering
\includegraphics[width=1\columnwidth]{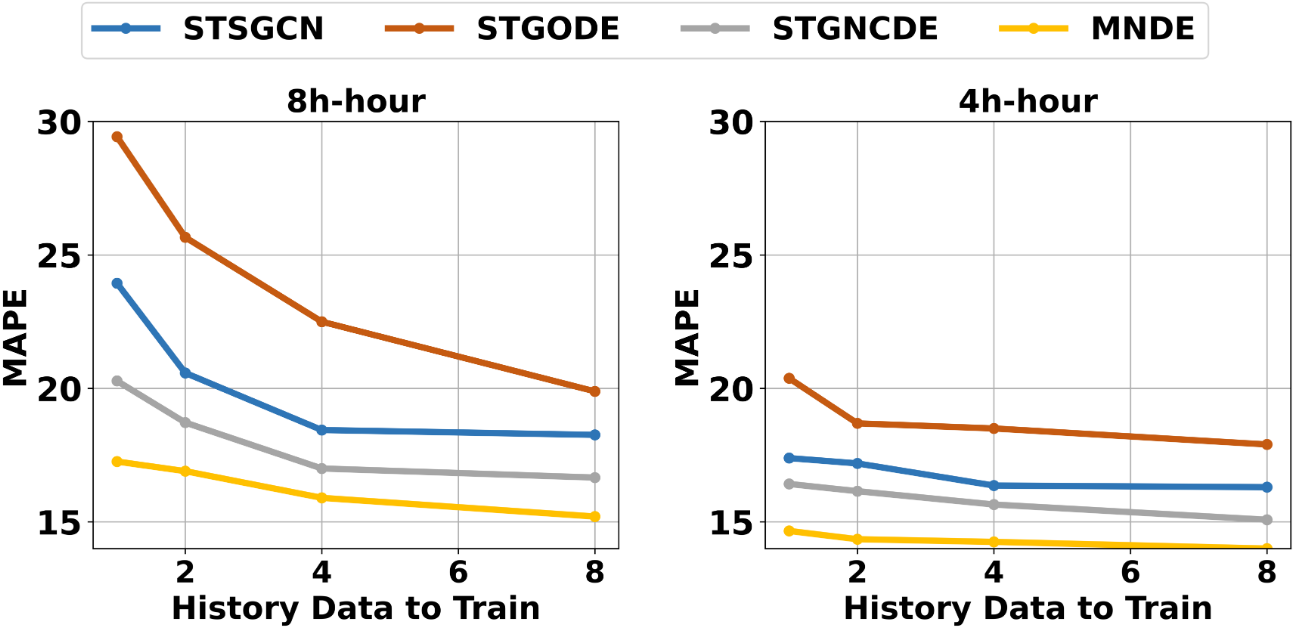}
% \captionsetup{font=small}
\caption{Performance comparison of four models with different length of input data on PEMS04 dataset. } 
\label{fig:length}
\end{figure}

We examine the performance of different models with varying lengths of input data ($l$). Figure \ref{fig:length} compares the proposed MNDE with STSGCN, STGODE and STGNCDE in terms of MAPE at the 8th hour time interval (the left plot) and the 4th hour time interval (the right plot) on PEMS04 dataset, with respect to the length of input data from 1 hour to 8 hours on the horizontal axis. STGODE, though being an NDE based method, exhibits relatively poor performance under different lengths of input data, likely due to the relative simplicity of its network architecture which captures limited traffic dynamics. STSGCN performs better but its non-NDE based model still cannot fully capture the traffic dynamics. STGNCDE, as an improved NDE based method, outperforms both STGODE and STSGCN. \modelname performs the best under all input lengths, thanks to its inclusion of delayed NDE, edge dynamics, and differentiation module, which are more powerful in capturing traffic dynamics.

\subsection{Delayed Pattern Visualization}

\begin{figure}[!t]
\centering
\includegraphics[width=\columnwidth]{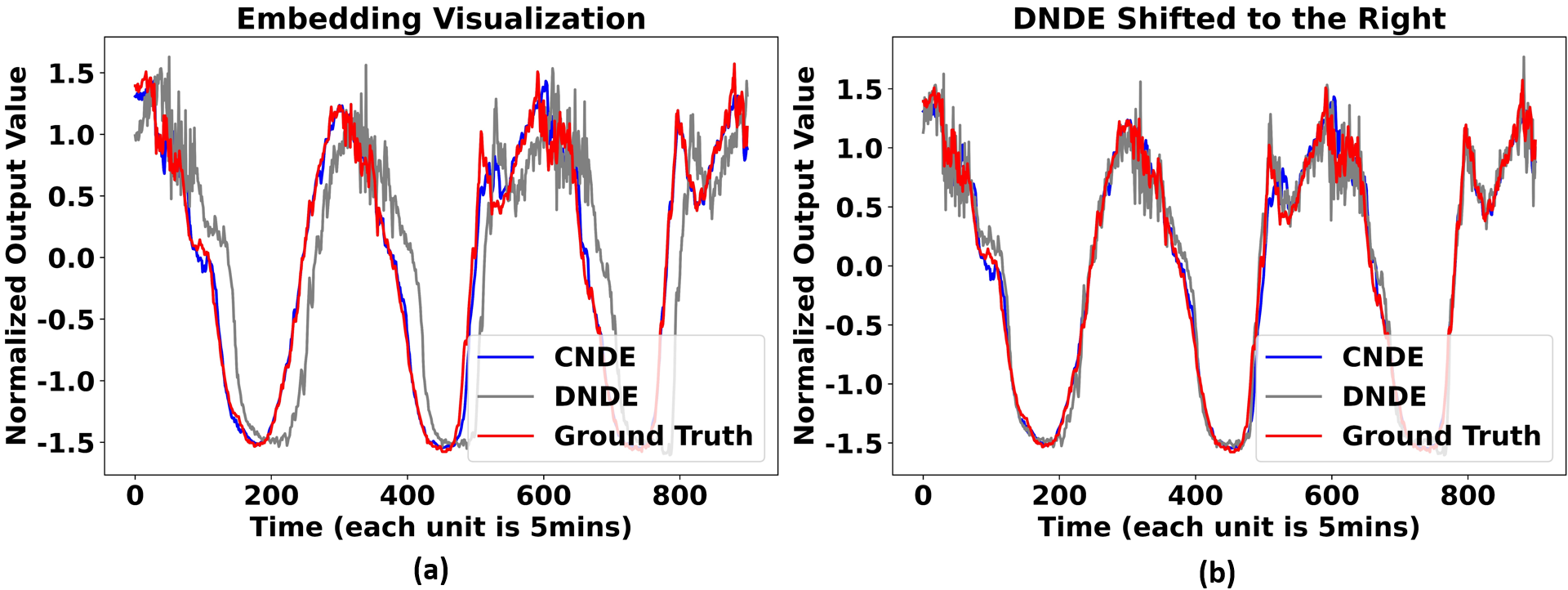}
\centering
\caption{(a) Embedding visualization. (b) Embedding visualization (DNDE shifted to the right)
} 
\label{fig:embedding}
\end{figure}

We demonstrate the impact of the DNDE and CNDE by presenting the 
Fig. \ref{fig:embedding}(a)-(b) visualized three variables, the ground truth, normalized aggregation of latent delay dynamics $H_{ST}^D(d)$ and $H_{E}^D(d)$ from DNDE, and normalized aggregation of latent current dynamics from  $H_{ST}^C(l)$ and $H_{E}^C(l)$ from CNDE. Fig. \ref{fig:embedding}(b) is the version after the variables shifting to the right. We can clearly discover the temporal shift in Fig. \ref{fig:embedding}(a), which proves the claims pointed out by DNDE Module, confirming the delayed dynamics play an important role in our model's overall superior performance.

\subsection{Results Under Different Flow Rate Ranges}
In Table \ref{tb:distribution}, we assign the ground truth of flow rate at each location and each time interval (5mins) into different flow rate ranges. And Table \ref{tb:distribution result} shows the prediction error of MNDE and STGNCDE within each range, where each prediction is produced by using one hour input data that is of 8 hours ago, for long-term forecasting. Our MNDE have a better performance in every flow rate range. 
\begin{table}[!h]
\centering
\caption{Flow rate distribution on PEMS08 dataset.}
\resizebox{\columnwidth}{!}{%
\begin{tabular}{|c|c|c|c|c|c|c|c|}
\hline
Flow Rate & 0-100 & 100-200 & 200-300 & 300-400 & 400-500 & 500-600 & \textgreater 600\\
\hline
count & 131580 & 144498 & 132829 & 93893 & 55238 & 24630 & 4161  \\
\hline
\end{tabular}
}
% \caption{Flow rate distribution.}
\label{tb:distribution}
\centering
\end{table}

\begin{table}[!h]
\centering
\centering
\caption{Forecasting error under different range of flow rate on PEMS08 dataset.}
\resizebox{\columnwidth}{!}{%
\begin{tabular}{|c|c|ccccccc|}
\hline
\multirow{2}{*}{\textbf{Metric}} & \multirow{2}{*}{\textbf{Model}} & \multicolumn{7}{c|}{\textbf{Flow Rate}} \\ \cline{3-9} 
                                & & \textbf{0-100} & \textbf{100-200} & \textbf{200-300} & \textbf{300-400} & \textbf{400-500} & \textbf{500-600} &\textbf{\textgreater600} \\ \hline
\multirow{2}{*}{MAE} & MNDE            & 14.72          & 22.64            & 27.27            & 31.14            & 34.55            & 37.18        & 55.74 \\
& STGNCDE          & 17.78          & 25.63            & 28.71            & 32.0             & 35.33            & 38.69        & 58.11 \\ \hline
\multirow{2}{*}{RMSE} & MNDE           & 25.95          & 34.62            & 39.75            & 44.82            & 49.61            & 55.66        & 74.94 \\
& STGNCDE         & 31.03          & 39.65            & 41.98            & 45.85            & 50.65            & 56.11         & 79.76\\ \hline
\multirow{2}{*}{MAPE(\%)} & MNDE           & 39.16          & 15.57            & 11.09            & 8.99             & 7.78             & 6.92          & 8.58\\
& STGNCDE         & 47.63          & 17.76            & 11.68            & 9.24             & 7.96             & 7.18          & 8.95\\ \hline
\end{tabular}
}

\label{tb:distribution result}
\end{table}

\subsection{Robustness With Zeros}
\begin{figure}[!t]
\centering
\includegraphics[width=\columnwidth]{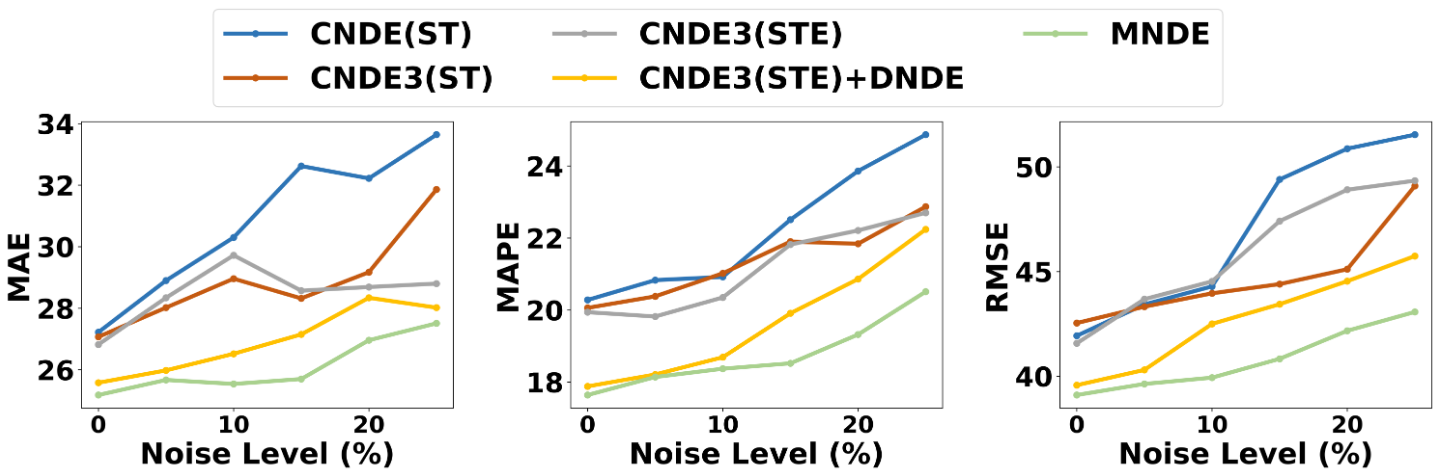}
\caption{ Robustness comparison of the models used in ablation study on PEMS08 dataset. }
\label{fig:robustness}
\end{figure}

We also conduct another type of robustness study by injecting noise into the input data. Different with the experiment shown in Table. \ref{tb:missing}, which could use Natural Cubis Spline to fill in ``NaN" by its neighbor's pattern, here we replaced the original values with zeros. ``NaNs" are the empty values but zeros are the real values taking into account for later model, which can potentially maximize uncertainties and noise that may arise during the data collection process. We varied the level of noise by adjusting the ratio of training data impacted by noise ($n$), ranging from $0\%$ (no noise) to $25\%$. Figure \ref{fig:robustness} presents the results of the robustness analysis, comparing the performance of the Single-NDE model and the models with additional components (+S, +E, +D) with \modelname on the PEMS08 dataset. Even when facing a high drop rate of $25\%$, \modelname outperforms many baseline models listed in Table 1. 

\subsection{Prediction Variance}

\begin{table}[ht]
\centering
\caption{Performance comparison on PEMS-BAY dataset. The \textbf{best} results are in bold and the  \underline{second-best} are underlined. }
\label{pems-bay}
\resizebox{1.0\columnwidth}{!}{%
\label{table:forecast_performance}
\begin{tabular}{|c|c|c|c|c|c|c|c|c|c|}
\hline
\multirow{2}{*}{\textbf{Model}} & \multicolumn{3}{c|}{\textbf{2nd hour}} & \multicolumn{3}{c|}{\textbf{4th hour}} & \multicolumn{3}{c|}{\textbf{8th hour}} \\ \cline{2-10}
& MAE & RMSE & MAPE & MAE & RMSE & MAPE & MAE & RMSE & MAPE \\ \hline
GraphWaveNet & 3.19 & 7.05 & 8.68 & 4.06 & 8.69 & 11.73 & 4.40 & 9.04 & 12.50 \\ \hline
STGCN & \underline{2.46} & 5.44 & 6.44 & 3.18 & 6.63 & 7.60 & 3.48 & 7.37 & 8.57 \\ \hline
STSGCN & 2.85 & 5.95 & 6.86 & 3.14 & 6.50 & 7.63 & 3.31 & 6.98 & 8.45 \\ \hline
STFGNN & 2.84 & 7.05 & 9.05 & 3.33 & 9.47 & 11.70 & 3.47 & 18.47 & 14.15\\ \hline
STGODE & 2.89 & 6.29 & 7.27 & 3.43 & 7.32 & 8.95 & 3.64 & 7.63 & 9.28 \\ \hline
MegaCRN & 2.89 & 7.89 & 6.66 & 3.95 & 11.26 & 8.81 & 4.64 & 12.86 & 9.59\\ \hline
FourierGNN & 3.64 & 7.52 & 9.67 & 4.57 & 9.14 & 12.74 & 4.83 & 9.66 & 13.27 \\\hline
ST-SSL & 2.68 & 5.65 & 6.53 & \underline{2.96} & 6.24 & 7.70 & 3.51 & 7.40 & 9.15 \\ \hline
STGNCDE & 2.49 & \underline{5.30} & \underline{6.05} & 2.99 & \underline{6.15} & \underline{7.34} & \underline{3.21} & \underline{6.94} & \underline{8.40} \\ \hline
MNDE & \textbf{2.42} & \textbf{5.27} & \textbf{5.96} & \textbf{2.69} & \textbf{5.75} & \textbf{6.89} & \textbf{3.09} & \textbf{6.54} & \textbf{7.83} \\ \hline
\end{tabular}
}
\end{table}

\begin{table*}[ht]
\centering
\caption{Performance comparison on METR-LA dataset. In each cell, mean is the left value, standard deviation is the value in the bracket. The \textbf{best} results are in bold and the  \underline{second-best} are underlined. }
\label{metr-la}
\resizebox{1.0\textwidth}{!}{%

\begin{tabular}{|c|c|c|c|c|c|c|c|c|c|}
\hline
\multirow{2}{*}{\textbf{Model}} & \multicolumn{3}{c|}{\textbf{2nd hour}} & \multicolumn{3}{c|}{\textbf{4th hour}} & \multicolumn{3}{c|}{\textbf{8th hour}} \\ \cline{2-10}
& MAE & RMSE & MAPE & MAE & RMSE & MAPE & MAE & RMSE & MAPE \\ \hline
GraphWaveNet & 6.34 (0.02) & 11.89 (0.05) & 22.00 (0.16) & 7.22 (0.02) & 13.42 (0.08) & 26.80 (0.16) & 7.46 (0.01) & 13.83 (0.04) & 27.76 (0.07) \\ \hline
STGCN & 8.91 (0.33) & 18.99 (0.30) & 17.57 (0.64) & 11.31 (1.23) & 22.02 (1.57) & 22.29 (2.40) & 13.07 (0.50) & 24.19 (0.78) & 25.76 (0.98) \\ \hline
STSGCN & \underline{4.98 (0.06)} & \underline{9.91 (0.16)} & 15.28 (0.22) & \underline{5.16 (0.10)} & \underline{10.32 (0.21)} & 16.60 (0.32) & \underline{5.39 (0.07)} & \underline{10.67 (0.07)} & 17.75 (0.64) \\ \hline
STFGNN & 11.83 (3.23) & 22.10 (3.50) & 22.93 (5.00) & 12.62 (2.47) & 23.30 (2.36) & 23.81 (4.27) & 13.40 (1.70) & 23.82 (1.80) & 25.26 (3.58) \\ \hline
STGODE & 6.54 (0.40) & 12.34 (1.06) & 20.81 (0.97) & 7.10 (0.64) & 13.24 (1.38) & 23.01 (1.14) & 6.98 (0.44) & 12.44 (0.63) & 21.89 (0.39) \\ \hline
MegaCRN & 10.35 (1.12) & 20.55 (1.25) & 21.20 (1.78) & 12.52 (1.22) & 23.60 (1.35) & 23.82 (1.98) & 14.43 (1.08) & 26.12 (1.18) & 25.45 (1.62) \\ \hline
FourierGNN & 7.16 (0.05) & 12.74 (0.02) & 21.54 (0.12) & 8.14 (0.10) & 14.30 (0.08) & 25.32 (0.08) & 8.75 (0.05) & 15.20 (0.12) & 26.75 (0.06) \\ \hline
ST-SSL & 5.50 (0.04) & 12.32 (0.06) & \underline{12.81 (0.28)} & 5.82 (0.02) & 12.86 (0.07) & 14.15 (0.32) & 6.46 (0.05) & 13.81 (0.03) & \underline{16.81 (0.13)} \\ \hline
STGNCDE & 5.10 (0.10) & 10.85 (0.25) & 13.01 (0.45) & 5.54 (0.11) & 11.85 (0.26) & \underline{13.98 (0.18)} & 6.29 (0.07) & 13.12 (0.18) & 16.96 (0.48) \\ \hline
MNDE & \textbf{4.87 (0.09)} &\textbf{ 9.23 (0.20) }& \textbf{12.58 (0.36)} & \textbf{5.01 (0.10)} & \textbf{10.00 (0.22)} & \textbf{13.05 (0.35)} & \textbf{5.25 (0.08)} & \textbf{10.33 (0.13)} & \textbf{16.04 (0.33)} \\ \hline
\end{tabular}
}
\end{table*}

Table \ref{pems-bay} presents the all model's accuracy performance on PEMS-BAY dataset. And Table \ref{metr-la} presents the average and standard deviation results of ten runs of our model using ten different random seeds. Results demonstrate that MNDE’s performance exhibits only minor variance across different seeds. Moreover, despite these variations, MNDE continues to outperform other baselines
across different benchmark datasets, as also evident by comparing to baseline results in Table \ref{tab:results}.

\subsection{Efficiency and Scalability}
\begin{figure}[!t]
\centering
\includegraphics[width=0.7\columnwidth]{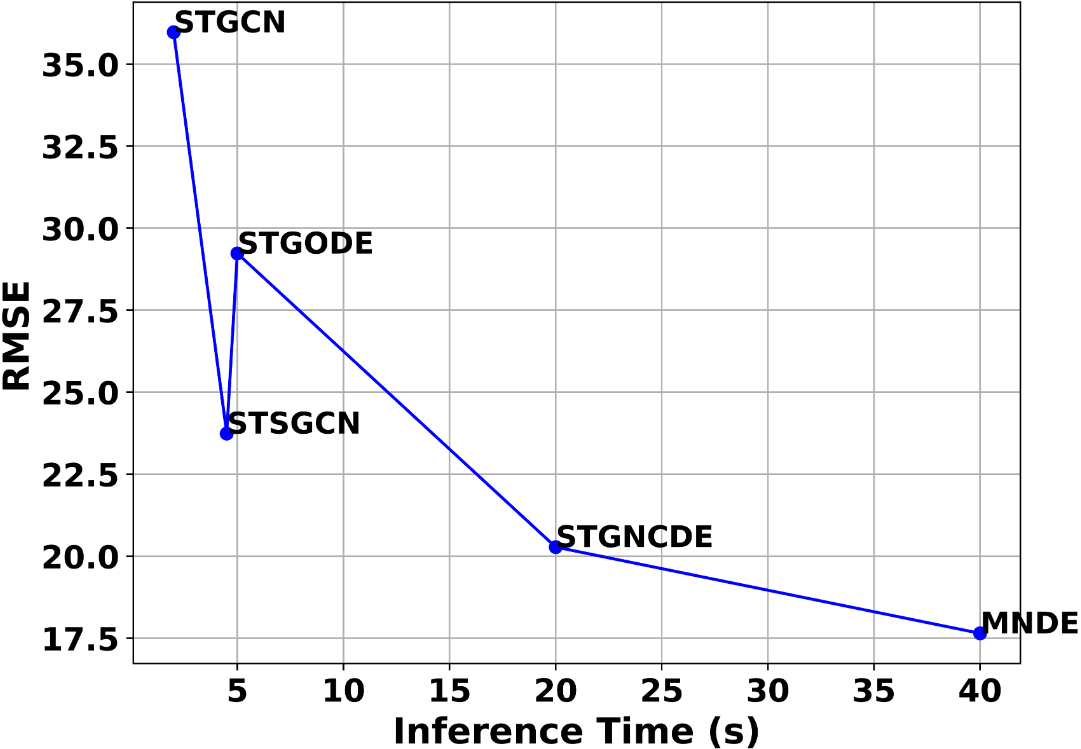}
\caption{ RMSE and inference time trade-off for various methods on the PEMS08 dataset. }
\label{fig:runtime}
\end{figure}

Fig. \ref{fig:runtime} illustrates the trade-off between RMSE forecasting performance and inference time for various methods. Although our MNDE framework exhibits a larger inference time, it delivers superior RMSE performance compared to other methods. This highlights that, in use cases where prediction accuracy is of paramount importance, the enhanced performance of our \modelname can outweigh the increased computational complexity. 

\begin{figure}[!t]
\centering
\includegraphics[width=0.7\columnwidth]{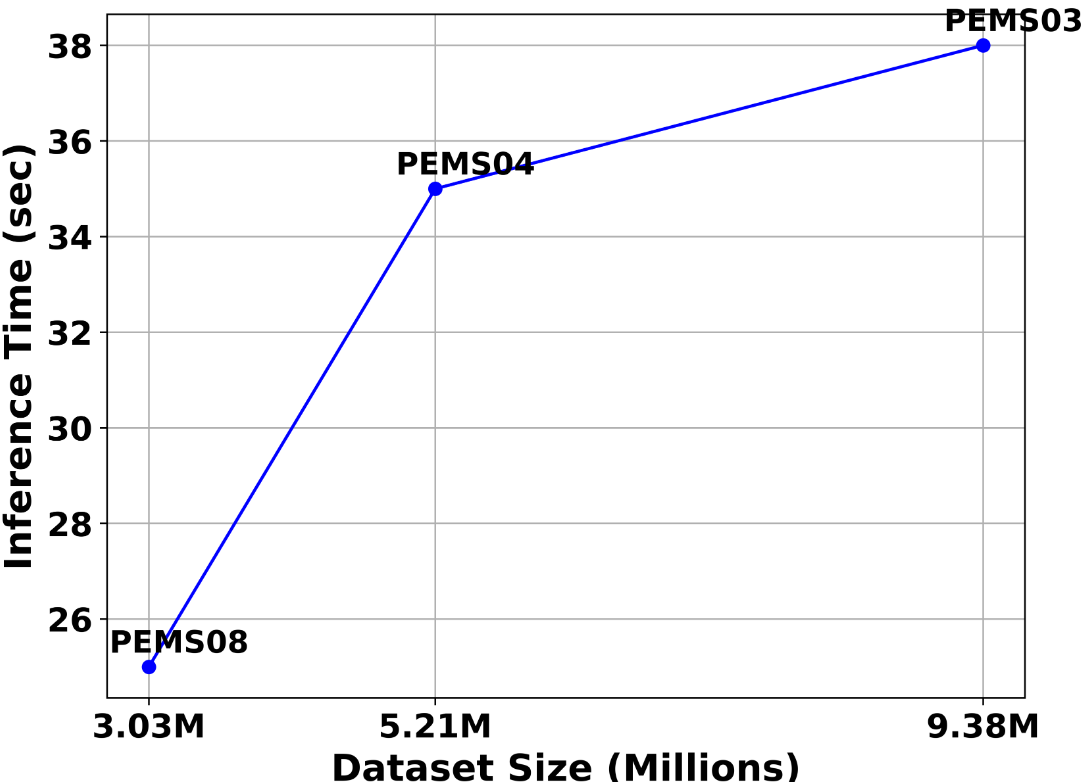}
\caption{  Inference time comparison by \modelname across datasets of varying sizes.  }
\label{fig:gap}
\end{figure}
Fig. \ref{fig:gap} showcases the inference time by \modelname. The dataset size on the x-axis is calculated by multiplying the number of nodes by the number of timesteps. As the dataset size increases, we observe that \modelname takes more time to inference. 

\vspace{-0.6em}
\section{Conclusion}
\label{sec:conclusion}
In this paper, we propose a new deep-learning framework, called Multi-View Neural Differential Equation (\modelname), for long-term spatio-temperal traffic flow forecasting. \modelname addresses several real-world challenges such as modeling delayed propagation of traffic conditions, capturing spatial dependency between locations explicitly, and handling local abrupt shift in traffic trend through a new  delayed neural differential equation module, a new dynamic edge dependency model, and a new differential module, respectively. \modelname uses multiple neural differential equations, which offer different views into the complex spatio-temporal dependencies in traffic flow data. Extensive experiments demonstrate the superior performance of our proposed model compared to other state-of-the-art models.
\section{Acknowledgement}
Shigang Chen’s work is supported in part by National Science Foundation under grant CNS-2312676 and National Institute of Health under grant R01 LM014027.

\section*{Acknowledgement}
Shigang Chen’s work is supported in part by the National Science Foundation under grant CNS-2312676 and the National Institute of Health under grant R01 LM014027. Zhe Jiang's work is supported by the National Science Foundation (NSF) under Grant No. IIS-2147908, IIS-2207072, and OAC-2152085.
\bibliographystyle{IEEEtran}

\bibliography{references}
\begin{IEEEbiography}[{\includegraphics[width=1in,height=1.25in,clip,keepaspectratio]{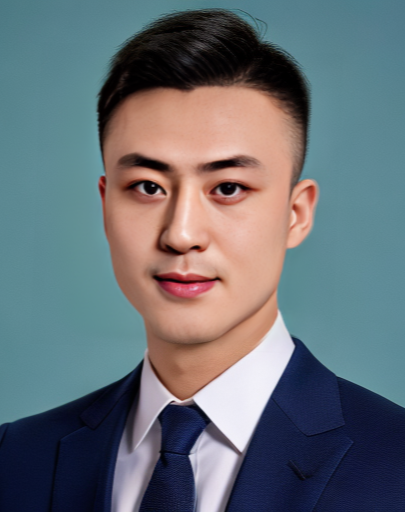}}]{Zibo Liu}
started his Ph.D. in Spring 2023 at the University of Florida. He is co-advised by Dr. Shigang Chen and Dr. Zhe Jiang. He received a B.S. in CS from Jilin University in China and Master's degree in CS from Virginia Tech. His research interests include data science, machine learning, and AI for computer networks, transportation, etc.
\end{IEEEbiography}
\begin{IEEEbiography}[{\includegraphics[width=1in,height=1.25in,clip,keepaspectratio]{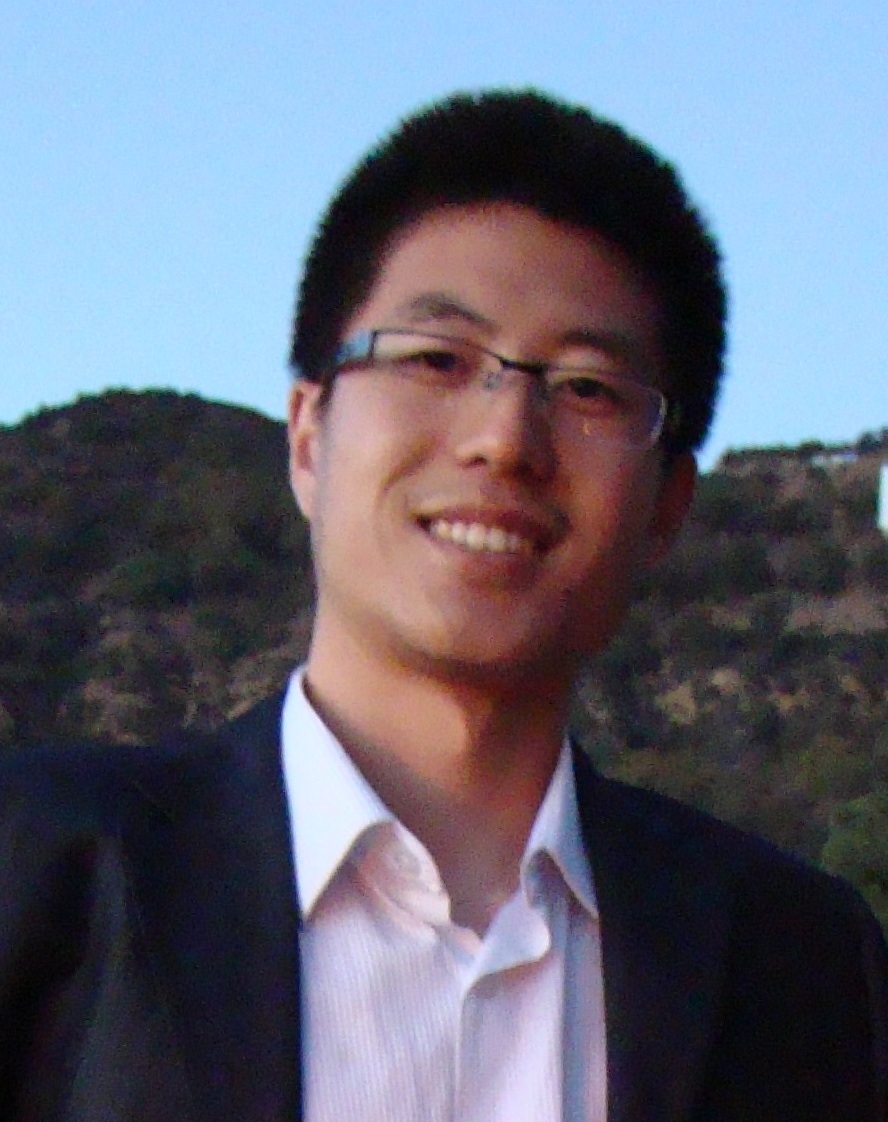}}]{Zhe Jiang}
(Senior Member, IEEE) received the
BE degree in electrical engineering and information science from the University of Science and
Technology of China, in 2010, and the PhD
degree in computer science from the University
of Minnesota, Twin Cities, in 2016. He is an
Assistant Professor with the Department of Computer Information Science \& Engineering, University of Florida. He is also affiliated with the
Center for Coastal Solutions. His research interests include data mining, deep learning, and spatiotemporal data mining, interdisciplinary applications in earth science, transportation, public health, etc.
\end{IEEEbiography}
\begin{IEEEbiography}[{\includegraphics[width=1in,height=1.25in,clip,keepaspectratio]{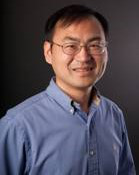}}]{Shigang Chen} (Fellow, IEEE) received the B.S. degree in computer science from the University of Science and Technology of China, China, in 1993, and the M.S. and Ph.D. degrees in computer science
from the University of Illinois at Urbana–Champaign
in 1996 and 1999, respectively. After graduation, he had worked with Cisco Systems for three years before joining the University of Florida in 2002. He is currently a Professor with the Department of Computer and Information Science and Engineering, University of Florida. He has
published more than 200 peer-reviewed journal/conference papers. He holds 13 U.S. patents, and many of them were used in software products. His research interests include computer networks, the Internet of Things, big data, cybersecurity, data privacy, edge-cloud computing, intelligent cyber-transportation systems, and wireless systems.
\end{IEEEbiography}

\end{document}